\documentclass[letterpaper, 10 pt, conference]{ieeeconf}  

\usepackage{graphicx}
\usepackage{mathtools}
\usepackage{bm}
\usepackage{algorithm}
\usepackage{algorithmic}
\usepackage{varwidth}

\newcommand{\matr}[1]{\bm{#1}}     

\IEEEoverridecommandlockouts                              

\title{\LARGE \bf
Surgical task-space optimisation of the CYCLOPS robotic system
}

\author{T.J.C. Oude Vrielink* $^{1}$, Y.W. Pang *$^{2}$, M. Zhao$^{1}$, S.-L. Lee$^{2}$, A. Darzi$^{1}$ and G.P. Mylonas, \textit{member,IEEE}$^{1}$
\thanks{*T.J.C. Oude Vrielink and Y.W. Pang are joint first authors}
\thanks{$^{1}$HARMS Lab, Department of Surgery, Imperial College London, W2 1PF London, UK. {\tt\small t.oude-vrielink15@imperial.ac.uk}}
	\thanks{$^{2}$ Department of Computing, Imperial College London}
}

\begin{document}

\maketitle
\thispagestyle{empty}
\pagestyle{empty}

\begin{abstract}
The CYCLOPS is a cable-driven parallel mechanism used for minimally invasive applications, with the ability to be customised to different surgical needs; allowing it to be made procedure- and patient-specific. For adequate optimisation, however, appropriate data on clinical constraints and task-space is required. Whereas the former can be provided through preoperative planning and imaging, the latter remains a problem, primarily for highly dexterous MIS systems. The current work focuses on the development of a task-space optimisation method for the CYCLOPS system and the development of a data collection method in a simulation environment for minimally invasive task-spaces. The same data collection method can be used for the development of other minimally invasive platforms. A case-study is used to illustrate the developed method for Endoscopic Submucosal Dissection (ESD). This paper shows that using this method, the system can be succesfully optimised for this application. 

\end{abstract}

\section{INTRODUCTION}
	\PARstart{M}{inimally} invasive surgery (MIS) has shown to be advantageous compared to open surgery in terms of postoperative pain, hospital stay and costs. In the last decades this has led to the introduction of laparoscopic surgery in clinical practice. To further expand MIS to other surgical domains, surgical systems are being developed aiming for natural orifice endoscopic surgery (NOES) and  natural orifice transluminal endoscopic surgery (NOTES). However, one major issue with current platforms is limited tissue triangulation, force delivery and control \cite{vitiello2013}. 

\begin{figure}
	\centering
	\includegraphics[width=3.5in]{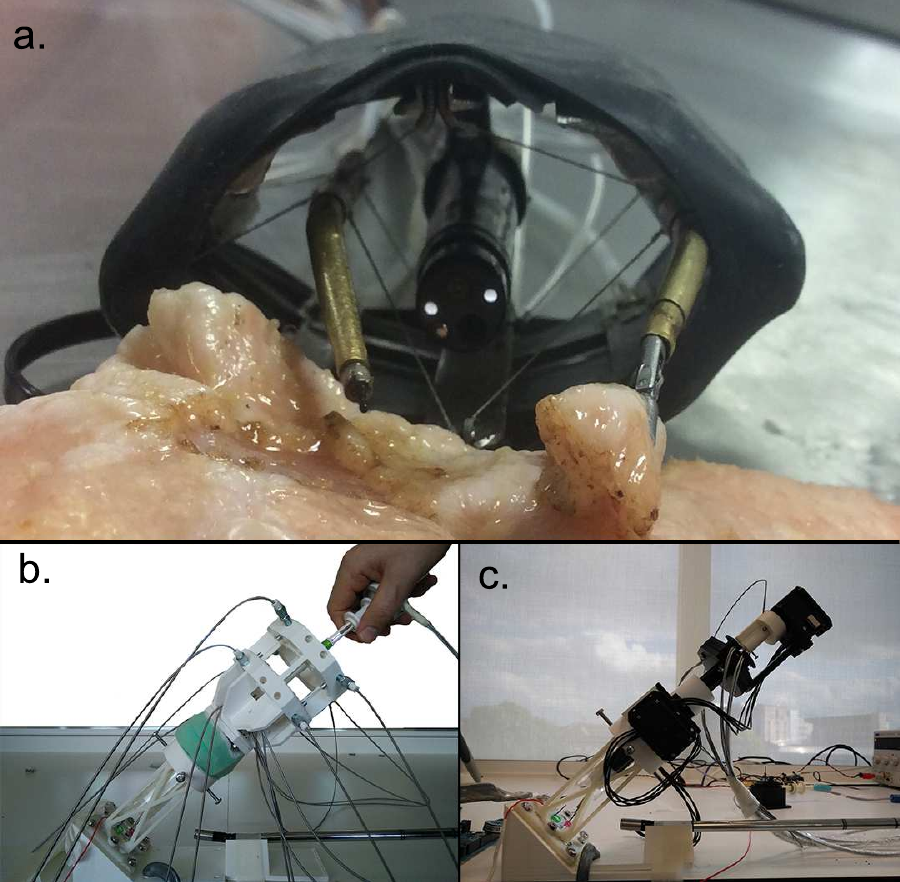}
	\caption{A) The latest version of the CYCLOPS, aiming for GI surgery and more specifically Endoscopic Submucosal Dissection. The system shown is currently undergoing extensive \textit{ex} and \textit{in vivo} animal trials. B) neuroCYCLOPS - A proof of concept of a hand-controlled CPDM for neurosurgical applications \cite{neuroCYCLOPS}. C) microCYCLOPS - A roboticed version of the neuroCYCLOPS which can be used for appications with cylindrical tissue retractors \cite{microCYCLOPS}. }
	\label{CYCLOPS}	
\end{figure}

The CYCLOPS is a surgical system that consists of a parallel tendon-driven mechanism for bi-manual control of surgical instruments. The latest version of the system, shown in Fig. \ref{CYCLOPS}a, is currently undergoing extensive pre-clinical \textit{in vivo} validation. In a first proof of concept, the platform has shown to exert forces on the end-effector of up to 65N, allowing for a large dexterous workspace and intuitive control \cite{c2014}. In addition, it was shown that the workspace is highly customizable by reconfiguring the attachment points on the surgical instruments and the entry points into the outer peripheral structure. The reconfigurability and customizability of cable-driven parallel mechanisms (CDPMs) for different workspace needs is one key advantage of CDPMs \cite{bostelman2000cable} \cite{pott2013ipanema}. As a result, the technical advantages the CYCLOPS brings to endoscopy can also be brought to other surgical applications. One variation of the CYCLOPS system is shown more recently by the development of the neuroCYCLOPS (\cite{neuroCYCLOPS}, Fig.\ref{CYCLOPS}b ) and microCYCLOPS (\cite{microCYCLOPS}, Fig.\ref{CYCLOPS}c).  A further range of applications is currently being explored, and to adequately design current and future systems, procedure-specific needs should be taken into account.

 
Current and near future developments in the fields of imaging, computation and manufacturing enable patient-specific optimisation of surgical systems. The combination of medical imaging modalities with advances in computational power is bringing advanced pre-operative planning to surgeries: estimation of the safe and improved access methods to reach the surgical site (e.g. in neurosurgery \cite{kassam2009completely}) and size of a specific cavity or instrument workspace after tissue retraction (e.g. pneumoperitoneum \cite{bano2012simulation}). At the same time, the improvement of manufacturing techniques (e.g. additive manufacturing) makes it economically feasible to create unique and customised parts, enabling patient-specific development of surgical instrumentation \cite{seneci2015rapid}. 

While technology enables the procedure-specific and patient-specific capability of the CYCLOPS platform, a limiting factor is the availability of workspace requirements for performing surgical tasks. In literature, one of the few systems developed for the collection of position and forces during a surgical tasks is the BlueDRAGON  \cite{rosen2002bluedragon}. This system has been used in surgery for the assessment of surgical skills, and establishment of design criteria for an optimised surgical system \cite{lum2006multidisciplinary}\cite{lum2006optimization}. While these data are interesting for laparoscopic surgery, the CYCLOPS is not constrained to the 4 degrees of freedom (DOF) of this application and can reintroduce the 6 DOF of open surgery to MIS applications \cite{vitiello2014augmented}. Studies focusing on instrument tracking during open surgical procedures \cite{d2016working}\cite{datta2001use}, are limited as the visualisation method and the instruments used influence the surgical task. In the absence of adequate taskspace data, additional data must be collected for the development of highly dexterous MIS platforms like the CYCLOPS. 

The current paper focuses on the optimisation of the CYCLOPS system to surgical taskspaces, and thereby enabling the procedure-specific and patient-specific capabilities of the platform. The optimisation method is discussed in the following section. In addition, a method was developed that enables the collection of taskspace data in a simulated surgical environment. This method is not limited to the CYCLOPS system and can be used for setting workspace requirements of other MIS systems. The final parts of the paper show a case-study in which the optimisation and data collection method is applied to Endoscopic Submucosal Dissection (ESD).

\section{OPTIMISATION METHOD}

The core concept of the CYCLOPS is shown in Fig \ref{fig_FBD}. The cables/tendons of the CDPM are introduced via fulcrum points into a rigid peripheral structure - the scaffold - and are connected to overtubes. Commercially available flexible instruments are placed into the overtube and manipulated by the tendons. The tendons are guided via pulleys or flexible transmission conduits - e.g. Bowden cables - to remotely placed motors. The current ESD CYCLOPS system, as shown in Fig \ref{CYCLOPS}a, uses additive manufacturing to create the stainless steel deployable scaffold. The design is parametric and can be optimised according to specific requirements, including anatomical constraints as well as taskspace requirements. The position of the tendon entry points into the scaffold, the  attachment points on the overtube, and the overtube shape deteremine the workspace of the instruments. Forces can only be exerted in the positive direction of the cables, and thus the number of tendons $ n $ must be at least one larger than the degrees of freedom $ m $: $ n \geq m + 1 $. The redundancy in actuation requires the calculation of the optimal tension distribution that satisfies the static or dynamic equilibria of the system. Section IIa introduces the method for calculation of the workspace and the optimal tension distribution. These concepts are used for the optimisation algorithm discussed in section IIb. Section IIc shows the optimisation algorithm compared to the standard configuration, as defined in that section. 
	
\subsection{Workspace estimation and optimal tension distribution}

\begin{figure}
	\centering
	\includegraphics[width=3.5in]{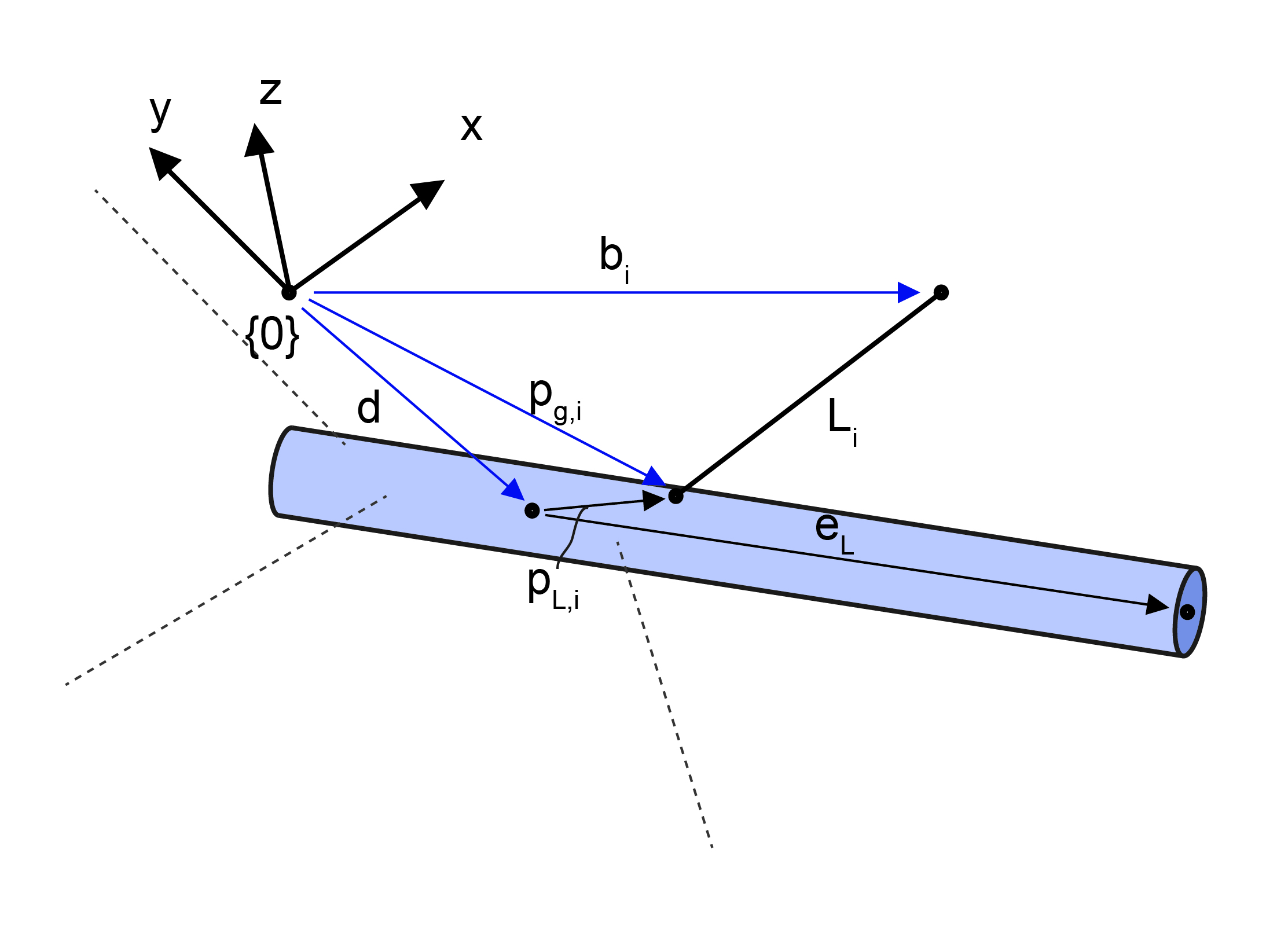}
	\caption{Convention used for describing the tool position, end-effector $e_l$ and tendon vectors $l_i$.}
	\label{fig_FBD}
\end{figure}

For the workspace estimation and the optimal tension distribution, the kinematics and the static equilibrium of the CYCLOPS system are required. The tendons, as shown in Fig. \ref{fig_FBD}, can be derived as vector $ \vec{l_i} $ from the attachment points on the overtube in the global coordinate framework $ \vec{p_{g,i}} $ to the feeding point into the scaffold $ b_i $:
\begin{equation}
 \vec{l_i} = \vec{b_i} - \vec{p_{g,i}} 
 \end{equation}
for $ i = 1,...,n$  tendons. The attachment point on the overtube depends on the instruments pose $ \vec{q} = [x,  y, z, \alpha, \beta, \gamma] $:
\begin{equation}
 \vec{p}_{g,i} = \matr{R}(\alpha,\beta,\gamma)\vec{P_{l,i}} + \begin{bmatrix}
 x  \\
 y  \\
 z
 \end{bmatrix}
 \label{eq_localtoglobal}
\end{equation}
Where $\matr{R}(\alpha,\beta,\gamma)$ represents the rotation matrix from the local tool coordinate system to the global scaffold coordinate system. 

 The static equilibrium of the instrument is described as:
\begin{equation}
\matr{A}\vec{t} + \vec{f} = 0
\label{eq_equilibrium}
\end{equation}
where $\vec{f}$ is the external wrench, $ \vec{t} $ the vector of tensions in the cables $t_i$ and $ \matr{A}$ is the structure matrix
\begin{equation}\matr{A} = \begin{bmatrix}
	\frac{l_1}{\lVert l_1 \lVert} && \dots && \frac{l_n}{\lVert l_n \lVert}\\
	p_{g,1}  \times \frac{l_1}{\lVert l_1 \lVert} && \dots && p_{g,n}  \times \frac{l_n}{\lVert l_n \lVert}
\end{bmatrix}
\label{eq_structMatrix}
\end{equation}

For manipulation of the instrument, the length of the tendons $ \lVert l_i \lVert $ is controlled. The Jacobian from the jointspace to the taskspace is the transpose of the structure matrix: $\matr{J} = \matr{A}^T $. Important to note are the trigonometric functions found in the rotation matrix  $\matr{R}$ (\ref{eq_localtoglobal}), making the Jacobian matrix non-linear. As a result, cable-driven tendon driven systems have straightforward inverse kinematics, though hard to compute forward kinematic. Consequently, the estimation of the workspace can be performed by conducting a gridsearch with a range of instrument poses. 

\subsubsection{Optimal Tension Distribution}
The redundancy in actuation compared to the degrees of freedom of the system make  (\ref{eq_equilibrium}) underconstraint. As a result, there are infinite possible tension vectors $ \vec{t}$ that satisfy this equation for a given pose. However, in reality there are physical constraints to the minimum and maximum tension the cables can take. Too much tension will result in failure of cables, whereas too low tension results in slackness and thus loss of controllability of the system. When calculating the workspace it is essential to only take those poses into account that satisfy the cable tension limits. 

There are different methods in literature to find the optimal tension distribution within predefined constraints. The minimum norm solution \cite{fattah2005design} has been shown to be computationally efficient, though has been shown to result in a too rough estimation of the workspace \cite{hay2005optimization}. More accurate solutions are the L1-norm and L2-norm method, and as the L1-norm can be computed efficiently using linear programming method, this is the preferred method. The L1-norm is described as the following optimisation problem:
\begin{equation}
	\underset{t}{\text{max}} \matr{1}^T \matr{t}
\end{equation}
subjected to
\begin{equation}
	\matr{A}\vec{t} = -\vec{f}
\end{equation}
\begin{equation}
t_{min} \leq t \leq t_{max}
\end{equation}
For a system with a single redundancy, the solution can be found analytically \cite{hay2005optimization}:
\begin{equation}
	\vec{t} = \begin{bmatrix}	-\matr{B}^{-1} \vec{h} \\ 1 \end{bmatrix} t_{redundant} + \begin{bmatrix}	-\matr{B}^{-1} \vec{f} \\ 1 \end{bmatrix}
\end{equation}
where $t_{redundant}$ is chosen as such that the resulting tension vector $ \vec{t} $ lies within the predetermined limits $ \{t_{min},t_{max}\}$ and 
\begin{equation}
\matr{A}^T = [\matr{B} \vec{h}]
\end{equation}
A more detailed explanation of this can be found in \cite{hay2005optimization}.

\subsection{Taskspace Optimisation algorithm}
The system workspace can be optimised to a specific taskspace by changing the size and shape of the overtube, tendon attachment and feeding points.

The taskspace is defined as the sequence of poses that the end-effector of the system assumes when a particular surgical procedure is being performed. Thus, a part of the objective function would involve trying to obtain a valid tension distribution for each point in the given sequence, causing the resulting objective function to be not continuous. This favours the use of algorithms based on particle swarm optimisation which was known to handle non-linear, discontinuous objective functions well \cite{perez2007particle}. 

A hybrid particle swarm pattern search algorithm \cite{vaz2009pswarm} is used to perform the optimisation. In addition, elements of two other optimisation methods, simulated annealing and particle filter optimisation (PFO), are incorporated to the algorithm.

Simulated annealing in global optimization involves using probability to accept or reject a new position in the search space. When incorporated to the algorithm, a probabilistic method is used to reject the current position of particle in the particle swarm and to re-sample it from the entire search space. This improved exploration of the parameter space.

Particle filter optimisation \cite{ji2008particle} is used at the end of the main algorithm. While particle filters are known to be less effective for high dimensionality problems as there are insufficient particles present to fully represent the distribution \cite{gustafsson2010particle}, it was noted that the particle swarm pattern search step causes a proportion of the particles to move towards areas which are more likely to be the global optimum. Re-sampling a new set of particles to these areas can help with exploration of these areas and provide a better result.

The final algorithm is outlined as follows:

\begin{algorithm}[H]
\caption{Optimization Algorithm to find Optimal Parameters for CYCLOPS given Task-space}
\begin{algorithmic} [1]
\renewcommand{\algorithmicrequire}{\textbf{Input:}}
 \renewcommand{\algorithmicensure}{\textbf{Output:}}
 \REQUIRE Task-space, Parameter bounds, CYCLOPS constants
 \ENSURE  Optimal CYCLOPS Parameters
  \STATE Initialize particle swarm by sampling uniformly from the parameter space
  \STATE Evaluate the objective function for each particle
 \\ \textit{Particle Swarm Pattern Search Loop}
  \FOR {$i <$ number of iterations}
  	\STATE Update velocities and positions of particles
  	\STATE Evaluate objective function values of particles 
  	\IF {no better position is found}
  		\STATE Perform pattern search poll step for best particle
        \IF {Poll step succeeds}
        	\STATE Increase pattern search mesh size
        \ELSE
        	\STATE Decrease pattern search mesh size
        \ENDIF
  	\ENDIF
  	\STATE Perform simulated annealing based random resampling mentioned above
  \ENDFOR
  \STATE Initialize particle swarm for PFO by sampling from previous population
 \\ \textit{Particle Filter Optimization Loop}
  \FOR {$i <$ number of PFO iterations}
  	\STATE Update velocities and positions of particles
    \STATE Evaluate objective function values of particles
    \STATE Update weights of particles
    \STATE Resample Particles based on weights to avoid degeneracy problem \cite{gustafsson2010particle}
  \ENDFOR
  \RETURN best position and best value
\end{algorithmic}
\end{algorithm}

\subsection{Comparison of Optimisation method with the standard configuration}

To illustrate the effect of the optimisation algorithm, a comparison is made with the standard configuration,  used as the tendon configuration in current CYCLOPS systems. The standard configuration is defined in this section.
The standard configuration has the tendons placed in an equilateral triangle (projected on the YZ-plane), as this has shown to be beneficial in terms of end-effector stiffness. The front and back entry points, as well as the front and back attachment points, are placed on a separate plane . The distance between the front and rear attachment point plane is defined as $ L_{att} $. Similarly, the distance between the front and rear entry points planes are defined as $ L_{entry} $. To fit a task-space within the workspace of the standard configuration, minor optimisation is done by a grid-search over parameters $ L_{att} $ and $ L_{entry}$. A full grid-search over all variables has shown to be cumbersome and computational expensive, and therefore, has never been used in practice.  

For comparison between both methods, a simple mock-up task was recorded with the data collection rig (discussed in next section), and the taskspace was placed in a convenient place for the standard configuration to reach all points in their predetermined orientations. Subsequently, our proposed optimisation method was set with the same size constraints.

\begin{figure}
	\centering
	\includegraphics[width=3.5in]{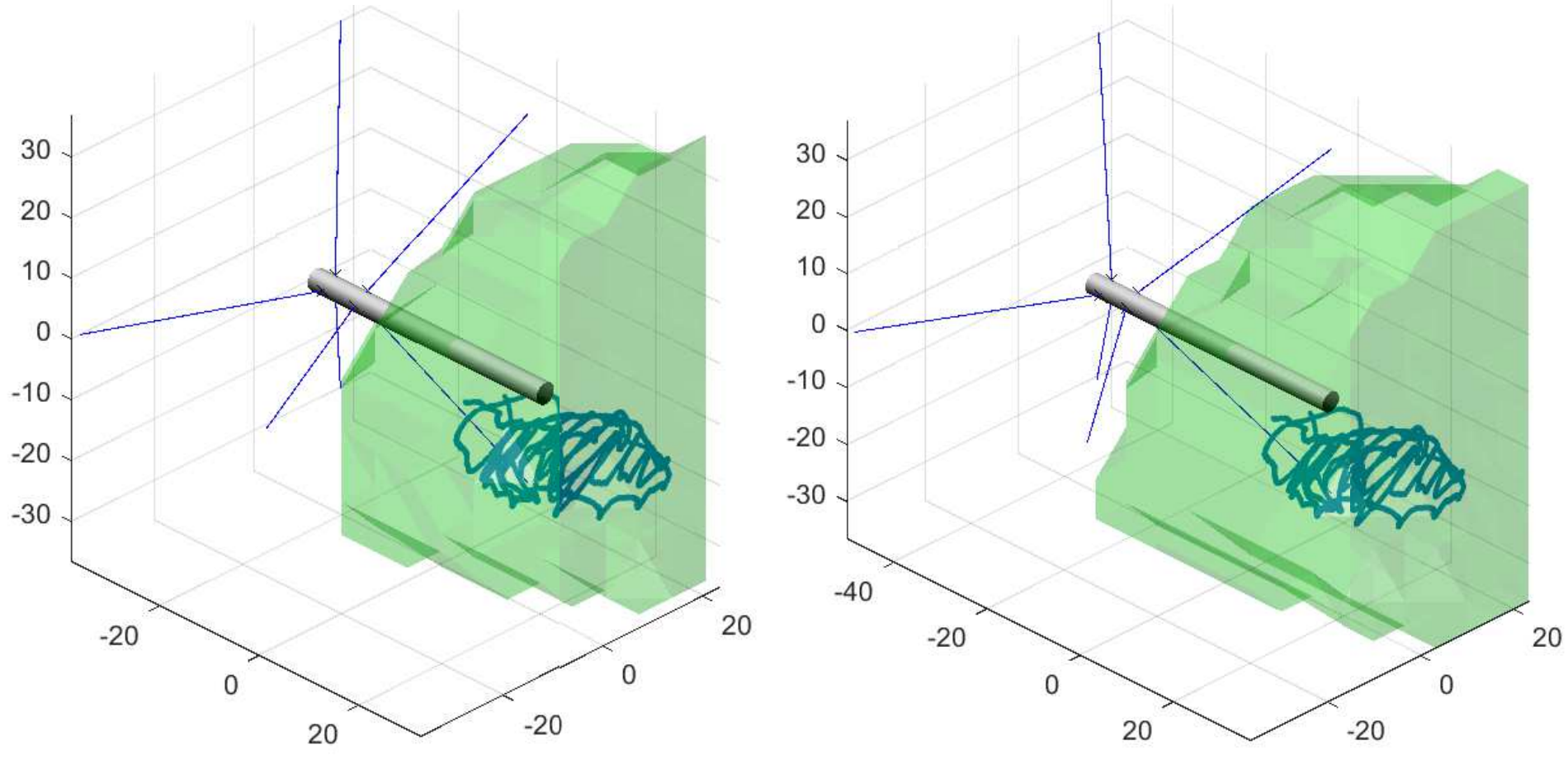}
	\caption{Baseline comparison between the standard configuration as defined in section IIc (left) and the optimisation method described in section IIb (right).  }
	\label{fig_baseline}	
\end{figure}

The results are shown in Fig. \ref{fig_baseline}. The volume of the workspace is calculated as $34.23cm^3$ and $48.18cm^3$, for the standard and optimised configuration, respectively. Both configurations can achieve the full taskpace. It can be seen that for this specific simple taskspace, the developed optimisation method is able to achieve a configuration that is able to increase the overal workspace of the system, without comprimising the task. A more extensive comparison is shown in section IV.  

\section{DATA COLLECTION}
	A data collection rig is created to obtain the taskspace used for optimisation, shown in Fig. \ref{fig_dataCollection}. The system uses an optical tracking system to track instruments performing the task. For the data to be realistic for endoscopic surgery, the task was performed through visualisation of the instruments through an endoscope and preventing direct visualisation to the surgeons hand.

\subsection{Optical Tracking Framework}
\begin{figure}
	\centering
	\includegraphics[width=3.5in]{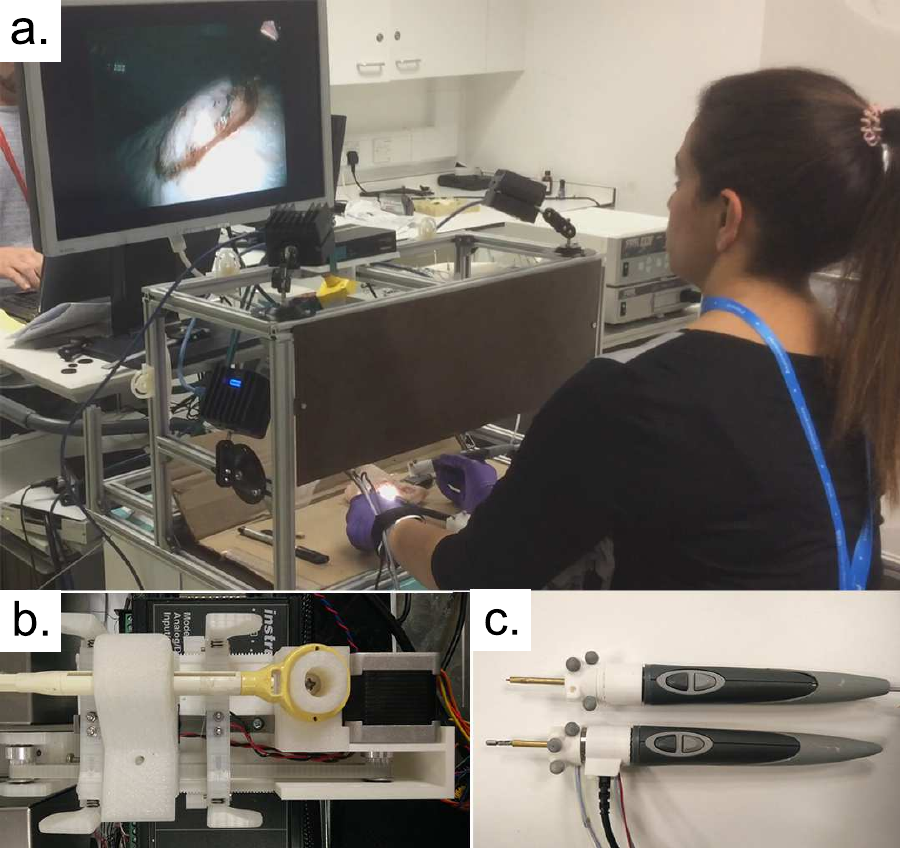}
	\caption{Setup for data collection of the simulated surgical task. \textbf{(a)}. The surgical task performed by a trained endoscopist. \textbf{(b)}. For the actuation of the flexible grasper, the actuation mechanism from the CYCLOPS system is used. \textbf{c}. The modified instruments with optical trackers and 6DOF loadcell.}
	\label{fig_dataCollection}
\end{figure}

For the tracking of the task-space of the instruments, 4 optical cameras (Optitrack Prime 13, NaturalPoint Inc, USA) are used, placed in an aluminium frame (Fig. \ref{fig_dataCollection}). The task is positioned in such a way to approximately place the initial position of the instruments in the centre of the optical tracking workspace. 

\subsection{Instruments}
Custom-made instruments are created for optical tracking, containing 3 optical markers (Fig. \ref{fig_dataCollection}c). Commercially available flexible instruments are used with motorised actuation for grasping and cutting. The instruments are placed within a rigid overtube that is in the longitudinal axis of the handle. The replaceable handles from the Geomagic Omni haptic manipulator are used to duplicate the ergonomics of the handle used for robotic surgery. The two buttons on these handles are used for actuation of grasping and cutting of the flexible instruments. External forces influence the wrench-feasible workspace and thus the tendon configuration. In particular during grasping and tissue lifting, will have to be taken into account. Therefore, a 6DOF loadcell (Nano 17, ATI Industrial Automation, USA) is added to flexible grasping instruments. The same instruments and actuation method are used in the CYCLOPS surgical system (Fig \ref{fig_dataCollection}b).

\section{CASE-STUDY: OPTIMISATION FOR ESD}
	This case-study focusses on the development and evaluation of the workspace design criteria for gastrointestinal surgery, and specifically aiming at an Endoscopic Submucosal Dissection (ESD), an intervention that is extremely difficult to perform minimally invasively, but with a high need for because of the high prevalence of flat adenomas worldwide. In ESD, depending on the area of interest, anatomical constraints vary from the less constraint gastric to the more constraint colorectal surgery. For this case-study we focus primarily on colorectal ESD surgery. 

\subsection{Anatomical considerations}
	
The colon diameter is an important anatomical constraint for the size of the CYCLOPS peripheral scaffold, and thus the tendon entry points. The size of the colon of 920 Japanese patients was measured, and depending on the location in the colonic tract, the diameter would vary from 33.2+-6.2mm in the Sigmoid to 49.1+-16.8mm in the Ascending colon \cite{sadahiro1992analysis}. A second study conducting analysis using CT scans estimated an 34.5+-7.1mm for the Sigmoid and 61.3+-11.1mm for the ascending bowel, however, the largest diameter was found to be the 75.7+-12.2mm for the Cecum \cite{khashab2009colorectal}. In addition to size, the colon tissue can be stretched up to 33\% an increase of mechanical stress or damage\cite{egorov2002mechanical}\cite{higa2007passive}. 

ESD is the preferred method of dissecting flat and depressed adenomas. The mean size of these flat lesions (an average from a group limited to all larger than 10mm in diameter) showed to be 20.8mm (n = 49) \cite{rembacken2000flat}.  

	
\subsection{Task Specification}
	The task is performed according to standard procedures, as described by Kakushima et al \cite{kakushima2008endoscopic}. This includes marking the lesion with a diathermy dissector, dissection around the lesion and dissection of the submucosa beneath the lesion. The instruments required for this task include a flexible needle, grasper and diathermy dissector. The left instrument is allocated for grasping (FG-44NR-1, Olympus, Japan), where the right is used for the diathermy (DualKnife KD-650L, Olympus, Japan) and needle. 
To simulate the task, a trained GI surgeon performed the ESD procedure on the skin of a chicken thigh. The area of interest is marked on the skin to be 20mm in diameter.

\subsection{Collected Data}	
\begin{figure}
	\centering
	\includegraphics[width=3.0in]{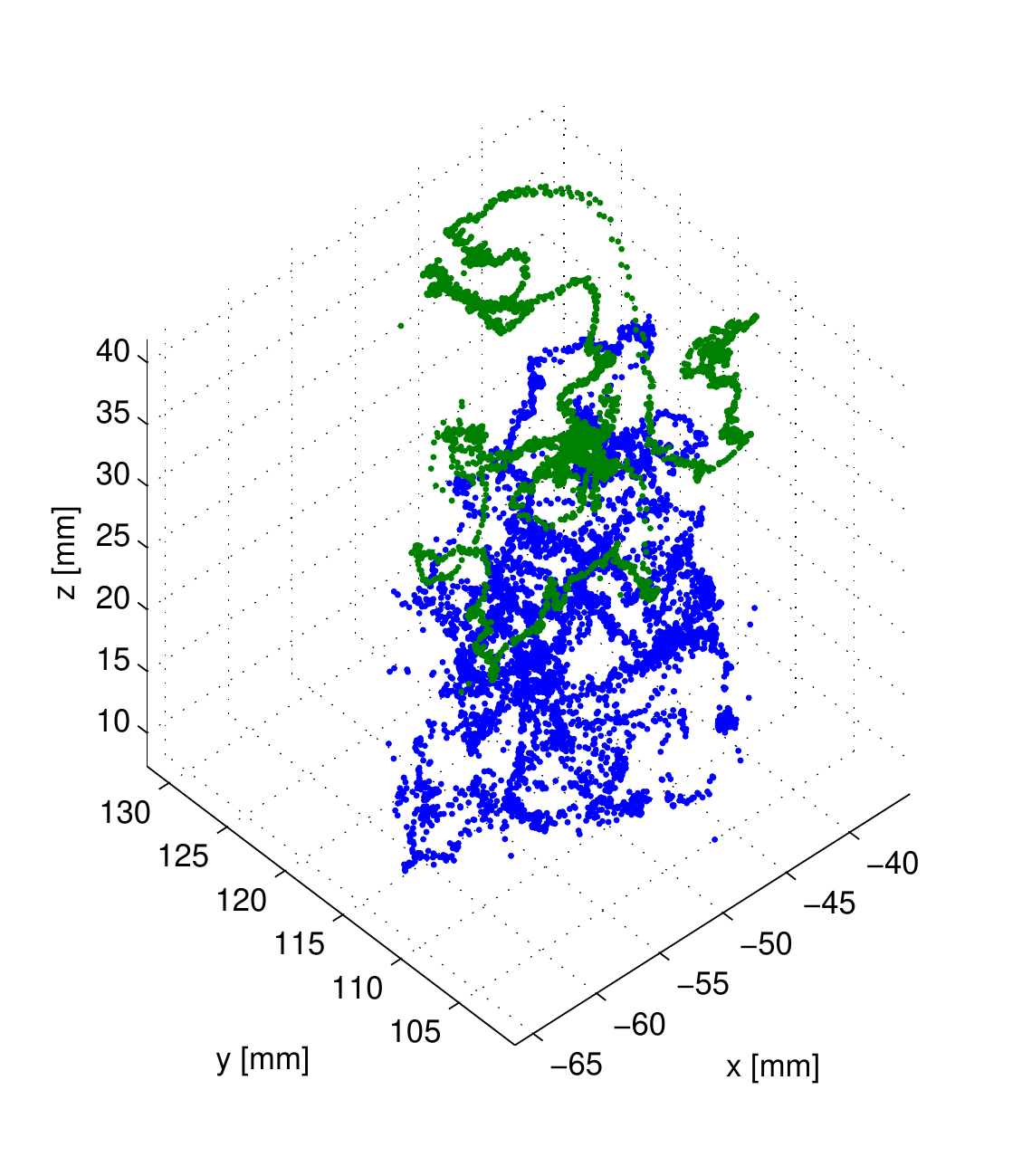}
	\caption{Taskspace of the two instruments. \textbf{Green}:grasping instrument (left hand), \textbf{Blue}: diathermy instrument (right hand). Endoscopic visualisation was provided in the YZ-plane, in positive X-direction. }
	\label{fig_3Dtaskspace}
\end{figure}

\begin{figure}
	\centering
	\includegraphics[width=3.5in]{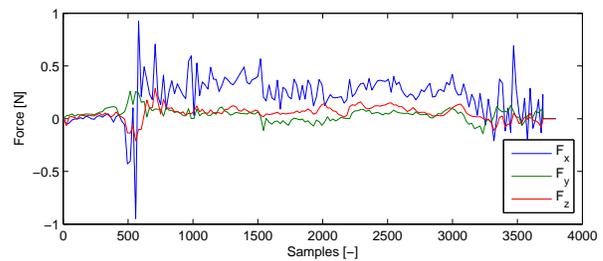}
	\caption{Measured forces of the left instrument, in the local frame of reference}
	\label{fig_forcesdata}
\end{figure}

	The data collected with the GI surgeon is shown in Fig. \ref{fig_3Dtaskspace}. The forces of the left instrument are shown in Fig. \ref{fig_forcesdata}. 

\subsection{Optimization Parameters and Procedure Constraints}
The optimization algorithm was constructed based on a 6-tendon CYCLOPS with a cylindrical scaffold. 

\subsubsection{Design Vector}

\begin{figure}
	\centering
	\includegraphics[width=3.5in]{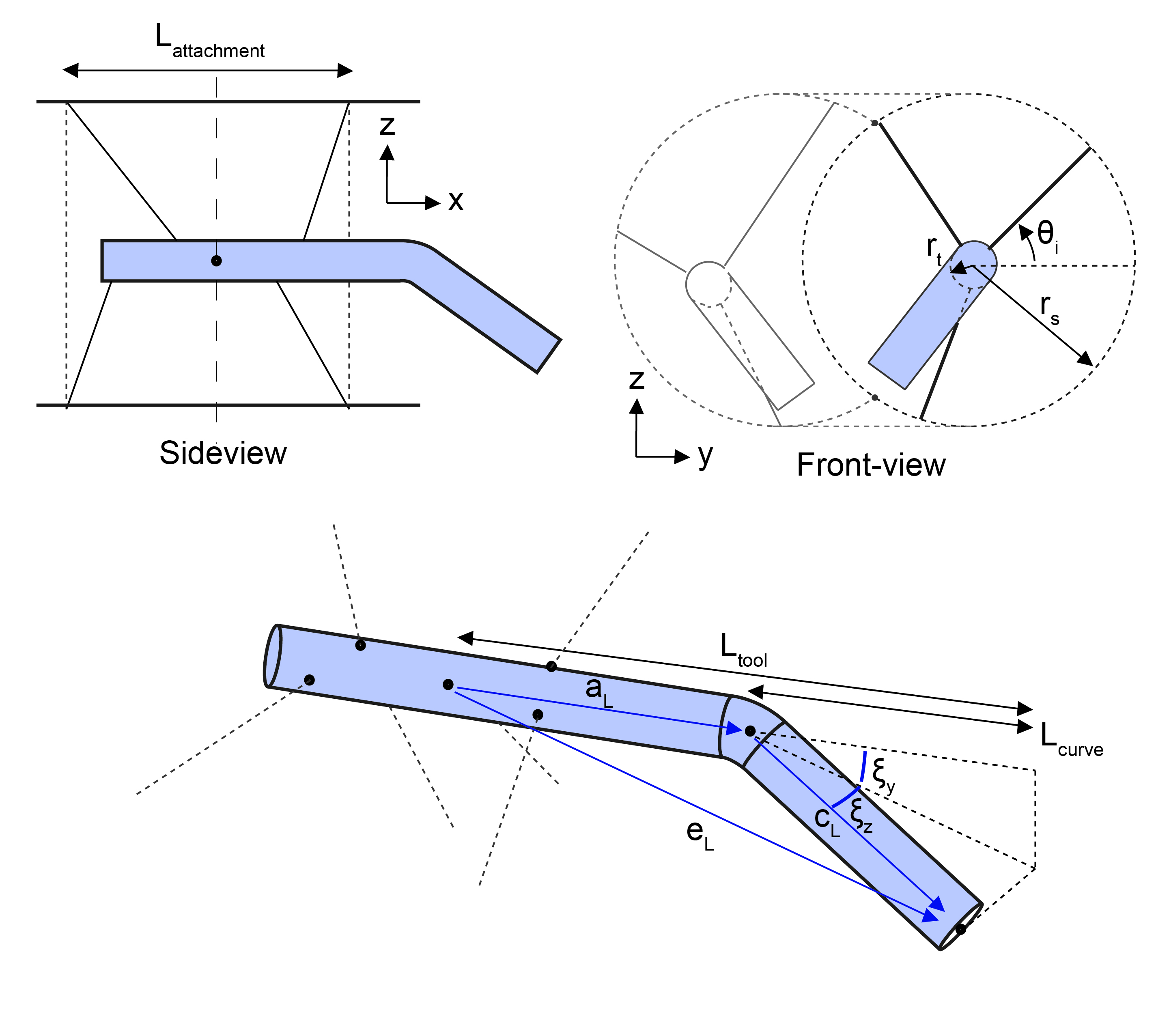}
	\caption{Schematics showing variables used for the design vector of the optimisation algorithm}
	\label{fig_REALSYSTEM}
\end{figure}

In the optimization algorithm, the following parameters are optimised (Fig. \ref{fig_REALSYSTEM}):
\begin{itemize}
\item Cross-sectional angle for each tendon, $\theta_i$
\item Attachment point for each tendon, $p_i$
\item Feeding point for front three tendons, $B_i, i \in \{1,2,3\}$
\item Feeding point for back three tendons, $B_i, i \in \{4,5,6\}$
\item Length of the tool, $L_{tool}$
\item Curvature angles, $\xi_y, \xi_z$, and curve position, $a_L$, if a curved tool is considered
\end{itemize}

If additional tendons are required, three parameters per additional tendon are added to the design vector: the cross sectional angle, the attachment point and the feeding point of each additional vector. The size of the design vector is thus given by $15 + 3(n-6) + 3c$, where n is the number of tendons, and $c \in \{ 0, 1 \}$ depending whether a curvature in the tool is required. 

\subsubsection{Objective Function}
The purpose of the optimization was to generate a CYCLOPS configuration that is superior to the standard configuration. This requires the objective function to return a suitable measure that can be optimized. In this study, the dexterous workspace defined as the set of positions that the system can achieve within a defined range of orientations \cite{hay2005optimization} was chosen. Such a measure aims to provide the surgeon with greater flexibility and maneuverability in controlling the tool. Additionally, some constraints were added due to practical considerations. The objective function therefore involves:

\newenvironment{renum}
  {\begin{enumerate}\renewcommand\labelenumi{\roman{enumi}.}}
  {\end{enumerate}}
\begin{renum}
\item Checking that tendons do not cross each other. This prevents tendon collisions during control of the CYCLOPS.
\item Ensuring that the $x$-distance between the feeding points is not smaller than that of the the attachment points. This is consistent with the present design of the CYCLOPS.
\item Checking that the input configuration is able to reach all points in the taskspace given.
\item Calculating the dexterous workspace
\end{renum}

\begin{figure}[htpb]
\centering
\includegraphics[width = 1\hsize]{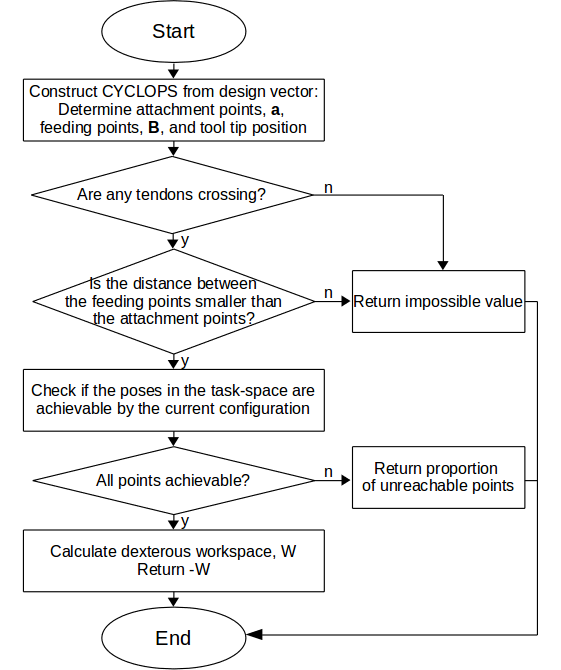}
\caption{Flowchart of Objective Function}
\label{fig:objfunction} 
\end{figure}

The objective function incorporates the above conditions and measures by returning values at different ranges depending on how well the input configuration satisfies the various constraints. Fig.~\ref{fig:objfunction} shows an outline of how this is done. \cite{bryson2016optimal} employs a similar method where the objective function returns a weighted sum of the condition and measure, with the condition having a much larger weight to ensure that it is fulfilled.

\subsection{Comparison Method}
The optimised system is compared to the standard configuration to illustrate the benefits. The standard configuration is defined similar as described in section IIc. To account for the curvature, the tip of the instrument is placed in the centre-point of the taskspace, and the mean yaw and pitch angles from the dataset is used to set the orientation. A comparison was maid for a bowel diameter of 70mm, for a single and double-curved instrument. The double-curved case was also compared for a 60mm bowel diameter. The minimal and maximum allowable tensions for both cases are set to $1N$ and $60N$,respectively. The forces on the left instrument are determined by the collected force data, whereas the right instrument has a weight of $ F_z = 0.1N$ acting on the overtube.
\section{Results}
\begin{figure*}
	\centering
	\includegraphics[width=\textwidth]{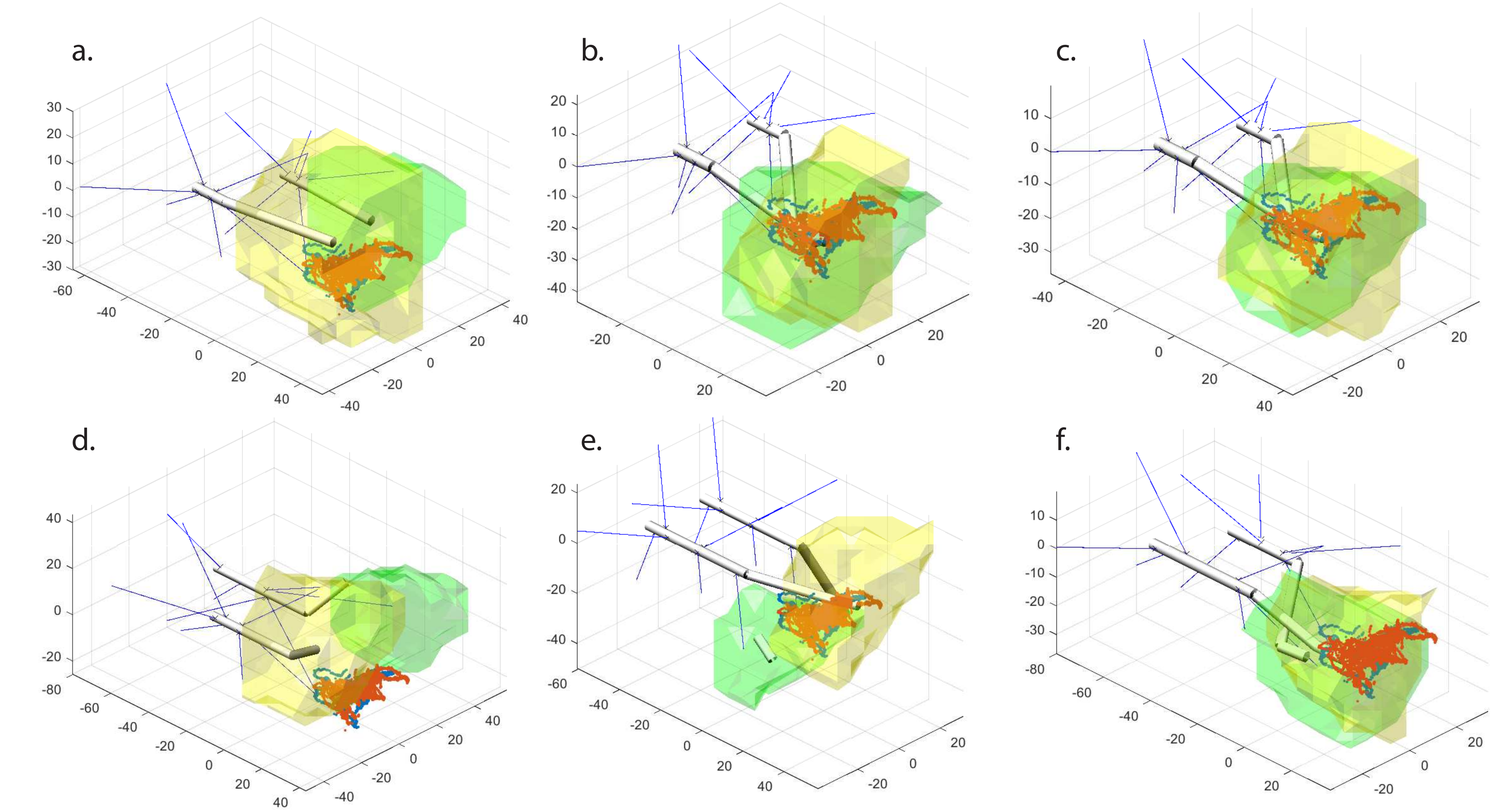}
	\caption{Found solutions for the standard configuration (top) and optimised configuration (bottom). \textbf{(a)} Standard configuration with single curvature, 70mm bowel diameter, \textbf{(b)} Standard configuration with double curvature, 70mm bowel diameter, \textbf{(c)} Standard configuration with double curvature, 60mm bowel diameter,\textbf{(d)} Optimised configuration with single curvature, 70mm bowel diameter, \textbf{(e)} Optimised configuration with double curvature, 70mm  bowel diameter, \textbf{(f)} Optimised configuration with double curvature, 60mm bowel diameter. Workspace size and percentage are shown in Tables \ref{table_right} and \ref{table_left}.}
	\label{fig_resCasestudy}	
\end{figure*}

\begin{figure}
	\centering
	\includegraphics[width=3.5in]{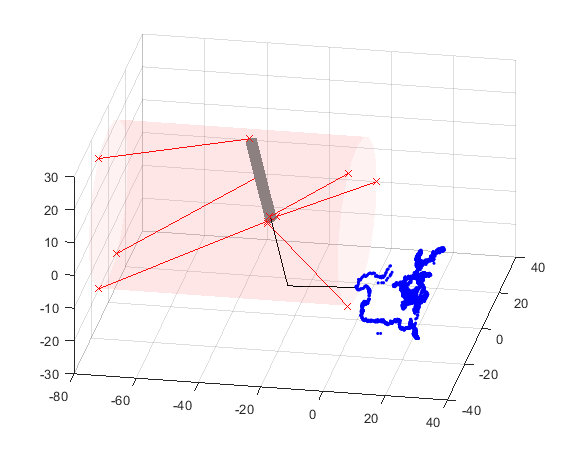}
	\caption{Optimised Configuration  - Single Curvature }
	\label{std_singCurv}	
\end{figure}

	The results are shown in Table \ref{table_right} and \ref{table_left}. The  corresponding configurations and zero-wrench reachable workspace are found in Fig. \ref{fig_resCasestudy}. In all cases the optimisation algorithm is able to get a higher percentage of the task-space, and smaller overall dexterous workspace. This indicates that the optimisation algorithm shows to be more efficient in customising the workspace to the required task-space. Unconventional results for the optimisation algorithm are most clear when we look at the left instrument of the single-curved solution. An extreme case is illustrated in Fig. \ref{std_singCurv}, from which it is easily seen that this is only in static equilibrium for a specific external force. In this specific moment, the external force on the tip of the instrument is $f_{ext} = [-1.6592,    0.0925,   -1.1588]^T N$.

\begin{table}
	\renewcommand{\arraystretch}{1.3}
	\caption{Right instrument }
	\label{table_right}
	\centering
	\begin{tabular}{r||ll|ll}
		\bfseries &  \bfseries Standard &  & \bfseries Optimised &\\	
			\bfseries &  \% task-space  &  $cm^2$ &  \% task-space  & $cm^2$ \\	
		\hline
		\hline
		   Single Curvature (70mm)		 & 48.93\% & 20.9	& 94.13\% &	9.44    \\
		   Double Curvature (70mm) 	 & 95.18\% & 15.13 	& 96.52\% &	4.65    \\
		   Double Curvature (60mm)  	 & 88.17\% & 11.25 	& 98.84\% & 7.51 	\\
	\end{tabular}
\end{table}

\begin{table}
	\renewcommand{\arraystretch}{1.3}
	\caption{Left instrument }
	\label{table_left}
	\centering
	\begin{tabular}{r||ll|ll}
		\bfseries &  \bfseries Standard &  & \bfseries Optimised &\\	
		\bfseries &  \% task-space  &  $cm^2$ &  \% task-space  & $cm^2$ \\	
		\hline
		\hline
		Single Curvature (70mm)	 & 28.35 \% & 15.01	& 100 \% &		2.59    \\
		Double Curvature (70mm) 	 & 96.26 \% & 13.72	& 100 \% &		5.28    \\
		Double Curvature (60mm)  	 & 81.50\% & 8.36 	& 89.74 \% &    7.34 	\\
	\end{tabular}
\end{table}

\section{Discussion}
	The baseline comparison, as shown in section IIc, shows that the optimisation algorithm is able to give a larger dexterous workspace when restricting the optimisation to a fixed scaffold length and diameter. This is achieved by a more optimal angle between tendons and attachment points on the overtube. The baseline comparison illustrates that for simple tasks, the optimisation algorithm is able to use the scaffold space more efficiently to achieve a large workspace. 

As shown, in the ESD case-study, the algorithm prioritises achieving the full task-space, rather than the size of the workspace. In the results, this is seen by the higher percentage of achieveable taskspace, but lower overall workspace when comparing to the standard configuration. For clinical practice, a larger workspace is beneficial in the ability to improvise for unforeseen circumstance during the surgical task, but a pre-requisite will be the ability to perform the main task, and therefore has priority. 

As this paper shows a proof of concept, currently only a single dataset is used. The found optimisation therefore cannot be generalised for all ESD procedures. The limitations of a single dataset become clear when forces are taken into account, resulting in poses that can only be achieved with these specific forces. By collecting data from multiple repeated tasks and over different surgeons, the data will become more representative for the taskspace that is required in clinical practice. The correct data handling for the optimisation algorithm for multiple datasets is essential part of future work for translation to clinical practice. 

Current paper only optimises for 70mm colon diameter and 60mm diameter. In reality, these values represent only larger sections of the bowel. It is expected that for smaller circumferences, the percentage of the achieveable taskspace will reduce further. However, in case of the right instrument, we can see that the 60mm diameter has better results than the 70mm diameter. To include such situations, the design vector can be changed to include the radius of the scaffold. Other ways to decrease the size of the system, while maintaining the ability to achieve the task is to add more degrees of freedom to the system - e.g. include roll or an articulated link on the overtube - or increase the number of tendons. The current approach focuses on representing the exact taskspace - including tool orientations - whereas in many cases the reachability of a point in space is more important than the orientation. Taking such aspects into account include different ways of handling the data (e.g weighing the position over the orientation), and are part of work required when dealing with multiple dataset, as mentioned in previous section.

\section{CONCLUSIONS}
	A first proof of concept data collection and optimisation method is shown in this paper. 
An initial baseline comparison has shown that the optimisation method is able to increase the dexterous workspace of the CYCLOPS for a given taskspace when similar constraints are set. In addition, the case-study shows that in case of realistic task and anatomical constraints, the optimisation algorithm will be able to achieve a higher percentage of the taskspace when compared to the standard optimisation method, and thereby fitting the workspace better to the required taskspace. Future work is focused on acquiring a larger dataset and validating the optimisation method for a physical CYCLOPS system.

\addtolength{\textheight}{-12cm}   




\section*{ACKNOWLEDGMENT}
This work is supported by the ERANDA Rothshild Foundation.

\bibliographystyle{IEEEtran}
\bibliography{bibfile}{}

\begin{thebibliography}{10}
\providecommand{\url}[1]{#1}
\csname url@samestyle\endcsname
\providecommand{\newblock}{\relax}
\providecommand{\bibinfo}[2]{#2}
\providecommand{\BIBentrySTDinterwordspacing}{\spaceskip=0pt\relax}
\providecommand{\BIBentryALTinterwordstretchfactor}{4}
\providecommand{\BIBentryALTinterwordspacing}{\spaceskip=\fontdimen2\font plus
\BIBentryALTinterwordstretchfactor\fontdimen3\font minus
  \fontdimen4\font\relax}
\providecommand{\BIBforeignlanguage}[2]{{%
\expandafter\ifx\csname l@#1\endcsname\relax
\typeout{** WARNING: IEEEtran.bst: No hyphenation pattern has been}%
\typeout{** loaded for the language `#1'. Using the pattern for}%
\typeout{** the default language instead.}%
\else
\language=\csname l@#1\endcsname
\fi
#2}}
\providecommand{\BIBdecl}{\relax}
\BIBdecl

\bibitem{vitiello2013}
V.~Vitiello, S.-L. Lee, T.~Cundy, and G.-Z. Yang, ``Emerging robotic platforms
  for minimally invasive surgery,'' \emph{Journal of Computational and Applied
  Mathematics}, vol.~6, pp. 111--126, 2013.

\bibitem{neuroCYCLOPS}
T.~Oude~Vrielink, D.~Khan, H.~Marcus, A.~Darzi, and G.~Mylonas, ``Neurocyclops:
  A novel system for endoscopic neurosurgery,'' in \emph{The Hamlyn Symposium
  on Medical Robotics}.\hskip 1em plus 0.5em minus 0.4em\relax HSMR, 2016.

\bibitem{microCYCLOPS}
T.~Oude~Vrielink, A.~Darzi, and G.~Mylonas, ``microcyclops: A robotic system
  for microsurgical applications,'' in \emph{6th Joint Workshop on New
  Technologies for Computer/Robot Assisted Surgery}.\hskip 1em plus 0.5em minus
  0.4em\relax CRAS2016, 2016.

\bibitem{c2014}
G.~Mylonas, V.~Vitiello, T.~Cundy, A.~Darzi, and G.-Z. Yang, ``Cyclops: A
  versatile robotic tool for bimanual single-access and natural-orifice
  endoscopic surgery,'' vol.~6.\hskip 1em plus 0.5em minus 0.4em\relax IEEE,
  2014, pp. 2436--2442.

\bibitem{bostelman2000cable}
R.~Bostelman, A.~Jacoff, F.~Proctor, T.~Kramer, and A.~Wavering, ``Cable-based
  reconfigurable machines for large scale manufacturing,'' in \emph{Proceedings
  of the 2000 Japan-USA Symposium on Flexible Automation}, 2000, pp. 23--26.

\bibitem{pott2013ipanema}
A.~Pott, H.~M{\"u}therich, W.~Kraus, V.~Schmidt, P.~Miermeister, and A.~Verl,
  ``Ipanema: a family of cable-driven parallel robots for industrial
  applications,'' in \emph{Cable-Driven Parallel Robots}.\hskip 1em plus 0.5em
  minus 0.4em\relax Springer, 2013, pp. 119--134.

\bibitem{kassam2009completely}
A.~B. Kassam, J.~A. Engh, A.~H. Mintz, and D.~M. Prevedello, ``Completely
  endoscopic resection of intraparenchymal brain tumors,'' \emph{Journal of
  neurosurgery}, vol. 110, no.~1, pp. 116--123, 2009.

\bibitem{bano2012simulation}
J.~Bano, A.~Hostettler, S.~Nicolau, S.~Cotin, C.~Doignon, H.~Wu, M.~Huang,
  L.~Soler, and J.~Marescaux, ``Simulation of pneumoperitoneum for laparoscopic
  surgery planning,'' in \emph{International Conference on Medical Image
  Computing and Computer-Assisted Intervention}.\hskip 1em plus 0.5em minus
  0.4em\relax Springer, 2012, pp. 91--98.

\bibitem{seneci2015rapid}
C.~A. Seneci, J.~Shang, A.~Darzi, and G.-Z. Yang, ``Rapid manufacturing with
  selective laser melting for robotic surgical tools: Design and process
  considerations,'' in \emph{Intelligent Robots and Systems (IROS), 2015
  IEEE/RSJ International Conference on}.\hskip 1em plus 0.5em minus 0.4em\relax
  IEEE, 2015, pp. 824--830.

\bibitem{rosen2002bluedragon}
J.~Rosen, J.~D. Brown, L.~Chang, M.~Barreca, M.~Sinanan, and B.~Hannaford,
  ``The bluedragon-a system for measuring the kinematics and dynamics of
  minimally invasive surgical tools in-vivo,'' in \emph{Robotics and
  Automation, 2002. Proceedings. ICRA'02. IEEE International Conference on},
  vol.~2.\hskip 1em plus 0.5em minus 0.4em\relax IEEE, 2002, pp. 1876--1881.

\bibitem{lum2006multidisciplinary}
M.~J. Lum, D.~Trimble, J.~Rosen, K.~Fodero, H.~H. King, G.~Sankaranarayanan,
  J.~Dosher, R.~Leuschke, B.~Martin-Anderson, M.~N. Sinanan \emph{et~al.},
  ``Multidisciplinary approach for developing a new minimally invasive surgical
  robotic system,'' in \emph{Biomedical Robotics and Biomechatronics, 2006.
  BioRob 2006. The First IEEE/RAS-EMBS International Conference on}.\hskip 1em
  plus 0.5em minus 0.4em\relax IEEE, 2006, pp. 841--846.

\bibitem{lum2006optimization}
M.~J. Lum, J.~Rosen, M.~N. Sinanan, and B.~Hannaford, ``Optimization of a
  spherical mechanism for a minimally invasive surgical robot: theoretical and
  experimental approaches,'' \emph{IEEE Transactions on Biomedical
  Engineering}, vol.~53, no.~7, pp. 1440--1445, 2006.

\bibitem{vitiello2014augmented}
V.~Vitiello, T.~Cundy, A.~Darzi, G.~Yang, and G.~Mylonas, ``Augmented
  instrument control for the cyclops robotic system,'' in \emph{The Hamlyn
  Symposium on Medical Robotics}, 2014, p.~29.

\bibitem{d2016working}
A.-L.~D. D'Angelo, D.~N. Rutherford, R.~D. Ray, S.~Laufer, A.~Mason, and C.~M.
  Pugh, ``Working volume: validity evidence for a motion-based metric of
  surgical efficiency,'' \emph{The American Journal of Surgery}, vol. 211,
  no.~2, pp. 445--450, 2016.

\bibitem{datta2001use}
V.~Datta, S.~Mackay, M.~Mandalia, and A.~Darzi, ``The use of electromagnetic
  motion tracking analysis to objectively measure open surgical skill in the
  laboratory-based model,'' \emph{Journal of the American College of Surgeons},
  vol. 193, no.~5, pp. 479--485, 2001.

\bibitem{fattah2005design}
A.~Fattah and S.~K. Agrawal, ``On the design of cable-suspended planar parallel
  robots,'' \emph{Journal of mechanical design}, vol. 127, no.~5, pp.
  1021--1028, 2005.

\bibitem{hay2005optimization}
A.~Hay and J.~Snyman, ``Optimization of a planar tendon-driven parallel
  manipulator for a maximal dextrous workspace,'' \emph{Engineering
  Optimization}, vol.~37, no.~3, pp. 217--236, 2005.

\bibitem{perez2007particle}
R.~E. Perez and K.~Behdinan, ``Particle swarm optimization in structural
  design,'' in \emph{Swarm Intelligence, Focus on Ant and Particle Swarm
  Optimization}.\hskip 1em plus 0.5em minus 0.4em\relax InTech, 2007.

\bibitem{vaz2009pswarm}
A.~I.~F. Vaz and L.~N. Vicente, ``Pswarm: a hybrid solver for linearly
  constrained global derivative-free optimization,'' \emph{Optimization Methods
  \& Software}, vol.~24, no. 4-5, pp. 669--685, 2009.

\bibitem{ji2008particle}
C.~Ji, Y.~Zhang, M.~Tong, and S.~Yang, ``Particle filter with swarm move for
  optimization,'' in \emph{International Conference on Parallel Problem Solving
  from Nature}.\hskip 1em plus 0.5em minus 0.4em\relax Springer, 2008, pp.
  909--918.

\bibitem{gustafsson2010particle}
F.~Gustafsson, ``Particle filter theory and practice with positioning
  applications,'' \emph{IEEE Aerospace and Electronic Systems Magazine},
  vol.~25, no.~7, pp. 53--82, 2010.

\bibitem{sadahiro1992analysis}
S.~Sadahiro, T.~Ohmura, Y.~Yamada, T.~Saito, and Y.~Taki, ``Analysis of length
  and surface area of each segment of the large intestine according to age, sex
  and physique,'' \emph{Surgical and Radiologic Anatomy}, vol.~14, no.~3, pp.
  251--257, 1992.

\bibitem{khashab2009colorectal}
M.~Khashab, P.~Pickhardt, D.~Kim, and D.~Rex, ``Colorectal anatomy in adults at
  computed tomography colonography: normal distribution and the effect of age,
  sex, and body mass index,'' \emph{Endoscopy}, vol.~41, no.~08, pp. 674--678,
  2009.

\bibitem{egorov2002mechanical}
V.~I. Egorov, I.~V. Schastlivtsev, E.~V. Prut, A.~O. Baranov, and R.~A.
  Turusov, ``Mechanical properties of the human gastrointestinal tract,''
  \emph{Journal of biomechanics}, vol.~35, no.~10, pp. 1417--1425, 2002.

\bibitem{higa2007passive}
M.~Higa, Y.~Luo, T.~Okuyama, T.~Takagi, Y.~Shiraishi, and T.~Yambe, ``Passive
  mechanical properties of large intestine under in vivo and in vitro
  compression,'' \emph{Medical engineering \& physics}, vol.~29, no.~8, pp.
  840--844, 2007.

\bibitem{rembacken2000flat}
B.~Rembacken, T.~Fujii, A.~Cairns, M.~Dixon, S.~Yoshida, D.~Chalmers, and
  A.~Axon, ``Flat and depressed colonic neoplasms: a prospective study of 1000
  colonoscopies in the uk,'' \emph{The Lancet}, vol. 355, no. 9211, pp.
  1211--1214, 2000.

\bibitem{kakushima2008endoscopic}
N.~Kakushima and M.~Fujishiro, ``Endoscopic submucosal dissection for
  gastrointestinal neoplasms,'' \emph{World Journal of Gastroenterology: WJG},
  vol.~14, no.~19, p. 2962, 2008.

\bibitem{bryson2016optimal}
J.~T. Bryson, X.~Jin, and S.~K. Agrawal, ``Optimal design of cable-driven
  manipulators using particle swarm optimization,'' \emph{Journal of mechanisms
  and robotics}, vol.~8, no.~4, p. 041003, 2016.

\end{thebibliography}




\end{document}